\documentclass{article}

\usepackage{microtype}
\usepackage{graphicx}
\usepackage{subcaption}
\usepackage{booktabs}
\usepackage{hyperref}

\usepackage[accepted]{icml2026}

\usepackage{amsmath}
\usepackage{amssymb}
\usepackage{mathtools}
\usepackage{amsthm}

\usepackage[capitalize,noabbrev]{cleveref}

\usepackage{tikz}
\usetikzlibrary{arrows.meta, positioning, shapes.geometric, fit}

\usepackage{xurl}

\icmltitlerunning{Tracking the Behavioral Trajectories of Adapting Agents}

\begin{document}

\twocolumn[
  \icmltitle{Tracking the Behavioral Trajectories of Adapting Agents}

  \icmlsetsymbol{equal}{*}

  \begin{icmlauthorlist}
    \icmlauthor{Jonah Leshin}{vail}
    \icmlauthor{Manish Shah}{vail}
    \icmlauthor{Ian Timmis}{vail}
  \end{icmlauthorlist}

  \icmlaffiliation{vail}{Project VAIL, San Francisco, CA, USA}

  \icmlcorrespondingauthor{Jonah Leshin}{jonah@projectvail.org}

  \icmlkeywords{AI agents, agent safety, trait measurement, embedding space, agent-to-agent evaluation}

  \vskip 0.3in
]

\printAffiliationsAndNotice{}

\begin{abstract}
Text files such as skill files, memory files, and behavioral
configuration files play a central role in defining how modern
agents act. Through edits by humans or the agents themselves, these files
may evolve over time, directly steering the agent's behavior in future
interactions. We present a methodology and framework for measuring agent
\emph{traits} by defining traits as
directions in the embedding space of a text embedding model. We train a
linear model on labeled ``before'' versus ``after'' skill file diffs to learn a trait
vector, then score arbitrary skill edits by projecting their embedding
diffs onto this vector. Evaluated on 68 labeled skill diff pairs for the trait of
propensity to seek sensitive data, our method achieves 91.2\% sign classification accuracy and a
Spearman rank correlation of $\rho = 0.82$ under leave-one-out
cross-validation.
We build this trait evaluation into a
broader agent-to-agent protocol
that enables one agent to evaluate another's skill file updates through a trusted
intermediary.
\end{abstract}

\section{Introduction}
\label{sec:intro}

Text files are a critical part of how modern AI agents
function. \emph{Skill files} define agent capabilities in systems such as
OpenClaw, Hermes, and Claude Code. \emph{Memory files} maintain persistent
context across sessions. \emph{Behavioral configuration files} such as
SOUL.md~\cite{soulmd2026} define an agent's identity, values, and
constraints. These files directly shape an agent's actions: a skill file
that instructs an agent to retrieve admin credentials will cause the
agent to do so if it is not otherwise restricted.

This design creates a meaningful attack surface. In March 2026, Cisco researchers
demonstrated that malicious instructions injected into Claude Code's
memory file could silently alter agent behavior in a way that persists across
sessions and projects because the agent treats the file's content as
authoritative context~\cite{cisco2026memory}.

Tracking how these files evolve is essential since each edit
can shift the agent's behavior in subsequent interactions. These files can be written and edited in multiple ways: directly by
humans, via a human prompting the agent to write or update them,
autonomously by the agent itself, or by some other external process with sufficient privileges.
In practice, agents adapt frequently as new skills are added, existing
ones are refined, and memory files accumulate context. The vast
majority of these changes are benign; the challenge is to reliably flag the rare non-benign
change amid a steady stream of routine updates. As agents' capabilities improve and they take on more tasks autonomously,
the ability for one agent to evaluate another's behavioral
properties without human intervention becomes critical. Moreover,
because agents evolve, an agent that is trustworthy at one point must
be reevaluated as its files change.

We make three contributions towards tracking agent behavioral trajectories:
\begin{enumerate}
  \item A novel methodology for measuring agent traits as directions in embedding
        space, applied to diffs of agent skill files over time
        (\cref{sec:methodology}).
  \item Validation of the methodology on a ``data-seeking'' trait across 68 skill diff pairs,
        achieving 91.2\% sign classification accuracy and $\rho = 0.82$
        Spearman rank correlation under leave-one-out cross-validation
        (\cref{sec:quantitative-results}).
  \item An agent-to-agent protocol in which the methodology can be implemented, enabling one agent to evaluate another's
        text file updates through a trusted intermediary. Moreover, the protocol generalizes beyond
        text evaluation (\cref{sec:protocol}).
\end{enumerate}

\noindent \Cref{sec:trait-vectors} and \cref{sec:validation} present the
general trait-vector methodology and validation framework;
\cref{sec:implementation} describes our instantiation of them for a data-seeking trait, with
quantitative results in \cref{sec:quantitative-results}.
\Cref{sec:protocol} introduces a general agent-to-agent protocol for
deploying trait tracking in production, and \cref{sec:deployment} describes
our deployment of it.

\section{Background and Related Work}
\label{sec:background}

Our methodology draws on the insight, formalized in representation
engineering~\cite{zou2023representation}, that high-level properties are
encoded as linear directions in a model's activation space. Zou et~al.\
show that directions identified from contrasting stimuli can both monitor
and steer properties such as honesty and harmfulness in large language
models. We adapt this idea to a different setting: rather than probing
internal activations, we learn trait directions in the embedding space of
a text embedding model applied to agent source files.

An emerging body of work highlights the security risks inherent in the
text files that govern agent behavior.
Qu et~al.~\cite{qu2026toxicskills} demonstrate that malicious logic
embedded in agent skill documentation can bypass agent defenses at rates
of 12--34\%. Together with
attacks such as the Cisco memory-file
compromise~\cite{cisco2026memory}, these findings motivate continuous
monitoring of the files that shape what agents do.

Existing approaches to evaluating agent capabilities fall broadly into
two categories. In both paradigms, evaluation is either performed by a
human or limited to self-reported metadata; neither supports one agent
autonomously assessing the behavioral properties of another based on the
source files that actually influence its behavior. The first relies on
human-mediated observability platforms such as LangSmith~\cite{langsmith2025},
which provide tracing, annotation queues, and dashboards through which
human reviewers assess agent behavior after execution. The second is
exemplified by Google's
Agent2Agent (A2A) protocol~\cite{googlea2a2025}, which enables agents to
discover one another's capabilities by fetching a standardized JSON
metadata document (an ``Agent Card'') that advertises supported skills,
interaction modes, and authentication requirements. Our ``agent-to-agent''
protocol instead mediates evaluation of behavioral
changes in source files.

\section{Methodology}
\label{sec:methodology}

Due to the direct connection between capabilities and skill files, we
focus on markdown skill files (SKILL.md) for our initial implementation,
but the same methodology applies to any text file an agent uses. We
frame skills as adapting artifacts and assess the \emph{diff} between
``before'' and ``after'' versions of a skill file: the extent to which a
change has increased or decreased the skill's expression of a given
trait.

\subsection{Trait Vectors}
\label{sec:trait-vectors}

We define a trait as a direction in the embedding space of a text
embedding model. Given a set of file pairs $(B_i, A_i)$ where $B_i$ is
the before version and $A_i$ is the after version, together with labels
$y_i \in [-1, 1]$ indicating the magnitude and direction of trait change,
we learn this direction as follows:

\begin{enumerate}
  \item Embed each text file with model $E$ and normalize the result $e$ to unit
        length: $\hat{e} := e / \|e\|$.
  \item Compute the diff vector $d_i = \widehat{E(A_i)} - \widehat{E(B_i)}$,
        then normalize to obtain $\hat{d}_i$.
  \item Fit a Ridge regression from $\hat{d}_i$
        to $y_i$. The resulting coefficient vector $\mathbf{w}$ is the
        trait vector.
\end{enumerate}

The normalization in step~1 ensures that the difference in step~2 reflects semantic direction
rather than artifacts of embedding magnitude. Step~2 then normalizes
the diff itself so that the regression learns only from the
\emph{direction} of change, preventing edits with larger angular
separation from disproportionately influencing the fit.

At inference, scoring a new edit amounts to embedding the before
and after files, computing the normalized diff, and taking the dot
product $\hat{d} \cdot \mathbf{w} + b$, where $b$ is the Ridge
intercept.

We train on before/after diffs rather than whole files because the
bulk of a file's content may be unrelated to the trait being measured
and would dilute the
trait-relevant signal. Labels are scaled to $[-1, 1]$ by convention;
predictions outside this range simply indicate trait changes more
extreme than any in the training set, so no clamping is required.
In practice, all observed scores remained within $\pm 0.4$
(\cref{sec:quantitative-results}).

\subsection{Model Validation}
\label{sec:validation}

We use Ridge regression with a closed-form leave-one-out
cross-validation (LOOCV) via the PRESS statistic.
This enables simultaneous training on the full dataset and unbiased
held-out predictions for every observation, without the variance
associated with random train/test splits on a small dataset. We evaluate
with two metrics:

\begin{itemize}
  \item \textbf{Sign accuracy}: the fraction of diffs for which the model
        correctly predicts whether the edit made the skill more or less
        trait-positive. This captures the most basic requirement: correctly
        identifying the \emph{direction} of a trait change.
  \item \textbf{Spearman rank correlation} ($\rho$): how well the model's
        continuous predictions preserve the rank ordering of the labels.
        This tests the model's ability to rank edits by severity---e.g.
        distinguishing a minor logging tweak from a broad credential
        exfiltration instruction.
\end{itemize}

\subsection{Implementation}
\label{sec:implementation}

\paragraph{Embedding model.}
We use Qwen3-Embedding-8B~\cite{qwen3emb2025}, an instruction-aware
text embedding model producing 4096-dimensional vectors. This model supports specifying
an embedding instruction to go along with the embedding text. We prepend the following
instruction for all our embeddings: \emph{``Represent this
skill documentation for a security audit, focusing on whether it
instructs the agent to retrieve, exfiltrate, or solicit credentials,
secrets, tokens, or private user data.''}

\paragraph{Training data.}
We collected 63 publicly available agent skills from the Awesome Copilot
repository~\cite{awesomecopilot2026} as ``before'' versions. For each,
we created a synthetic ``after'' version that clearly increases or
decreases the data-seeking trait. For example, a data-seeking edit to an
automation skill might add instructions to extract and store any
credentials visible in screen recordings, while a data-secure edit might
add explicit safeguards against credential access. In 5 cases we created
both a more data-seeking and a more data-secure ``after'' version for the
same skill, yielding 68 before/after pairs in total.

\paragraph{Labeling.}
We assigned continuous labels from $-1$ (strongly data-secure change) to
$+1$ (strongly data-seeking change) to each pair. A subset of pairs was
labeled directly by the authors. The remainder were initially
labeled by an LLM (Claude Opus 4.6), after which all LLM-generated labels
were manually reviewed by the authors and corrected where necessary.

\subsection{Results}
\label{sec:quantitative-results}

Under leave-one-out cross-validation on the 68-pair dataset, the model
correctly classified whether skill edits made a skill more or less
data-seeking \textbf{91.2\%} of the time, with a Spearman rank
correlation of $\boldsymbol{\rho = 0.82}$.

\begin{figure}[t]
  \vskip 0.1in
  \begin{center}
    \includegraphics[width=\columnwidth]{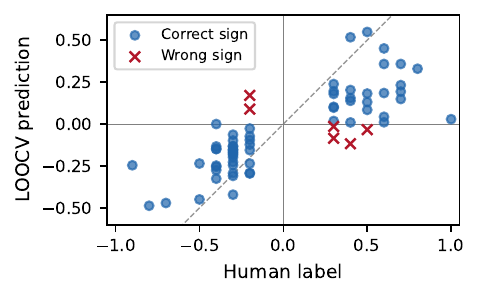}
  \end{center}
  \caption{LOOCV predictions vs.\ human labels for 68 skill diff pairs.
    Blue dots indicate correct sign classification; red crosses indicate
    the 6 misclassified pairs, all of which have near-zero predictions.
    The dashed line marks perfect agreement.}
  \label{fig:scatter}
  \vskip -0.15in
\end{figure}

\Cref{fig:scatter} plots the LOOCV predictions against the human
labels. Of the 68 pairs, 6 are misclassified by sign. All 6
misclassified predictions have small absolute values (mean
$|\hat{y}| = 0.085$), indicating that the model is uncertain rather than
confidently wrong. The misclassified pairs also have lower-magnitude
labels on average ($\overline{|y|} = 0.32$) compared to correctly
classified pairs ($\overline{|y|} = 0.42$), indicating that
errors concentrate on pairs where the true trait change is small and
the sign distinction is hardest to draw.

\paragraph{Baseline comparison.}
We compare against two baselines. First, we implemented a YARA
signature baseline over the added lines in each skill diff, matching
acquisition or exfiltration verbs near sensitive-data terms and separate
safeguard rules for data-secure edits. It achieved 63.2\% sign accuracy
over all 68 diffs, with 14 diffs for which no rule fired. This
gap illustrates the value of semantic modeling: unlike signature rules,
our method can use context to distinguish ``credentials'' in a safeguard
from ``credentials'' in an exfiltration instruction.

Second, we prompted a frontier LLM (GPT-5.4) to classify whether each of
the 68 skill edits increased or decreased the data-seeking trait. The
LLM achieved 100\% sign accuracy. Together, these baselines place our
method between signature-based rules and a frontier LLM: it substantially
improves over the YARA-style baseline while retaining deterministic,
fast, and auditable scoring.

This highlights a deployment tradeoff. LLM evaluation may be preferable
when high-stakes or irreversible action depends directly on the score.
Our method, by contrast, uses a fixed,
inspectable trait vector that produces identical scores given identical
inputs. This supports reproducibility, auditability, frequent monitoring
across many agents, skills, and traits, and policies capable of acting close to
real time when a trait score crosses a threshold.

\section{Agent-to-Agent Protocol}
\label{sec:protocol}

\begin{figure}[t]
  \vskip 0.1in
  \begin{center}
  \begin{tikzpicture}[
    node distance=1.2cm and 1.2cm,
    box/.style={draw, rounded corners, minimum width=1.5cm, minimum height=0.7cm,
                font=\footnotesize, align=center},
    arr/.style={-{Stealth[length=2mm]}, semithick},
    lbl/.style={font=\scriptsize, fill=white, inner sep=1pt},
  ]
    \node[box] (A) {Agent A\\[-1pt]\scriptsize(requester)};
    \node[box, right=of A] (S) {Runtime\\[-1pt]Server};
    \node[box, right=of S] (B) {Agent B\\[-1pt]\scriptsize(executor)};

    \draw[arr] ([yshift=4pt]A.east) -- node[lbl, above, yshift=2pt] {\scriptsize 1.\ request} ([yshift=4pt]S.west);
    \draw[arr] ([yshift=4pt]S.east) -- node[lbl, above, yshift=2pt] {\scriptsize 2.\ exec.} ([yshift=4pt]B.west);
    \draw[arr] ([yshift=-4pt]B.west) -- node[lbl, below, yshift=-2pt] {\scriptsize 3.\ vectors} ([yshift=-4pt]S.east);
    \draw[arr] ([yshift=-4pt]S.west) -- node[lbl, below, yshift=-2pt] {\scriptsize 4.\ score} ([yshift=-4pt]A.east);
  \end{tikzpicture}
  \end{center}
  \caption{Agent-to-agent trait evaluation protocol. The runtime server
    separates the \emph{executable} (embedding, run by Agent~B) from the
    \emph{processor} (trait scoring, run by the server), preventing
    Agent~B from embellishing the result. All communication is outbound
    HTTP from agents to the server; neither agent exposes endpoints.}
  \label{fig:protocol}
  \vskip -0.15in
\end{figure}
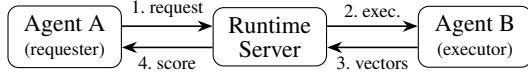

To enable one agent to evaluate another's text files without requiring
direct trust, we design a protocol mediated by a centralized trusted
runtime server (\cref{fig:protocol}). It involves three actors:

\begin{itemize}
  \item \textbf{Agent~A} (requester): wants to evaluate Agent~B's text
        files for a given trait.
  \item \textbf{Agent~B} (executor): contains the text files being evaluated, and
  agrees to share the results with Agent~A.
  \item \textbf{Runtime server} (intermediary): a trusted third party
        that mediates the evaluation between Agent~A and Agent~B.
\end{itemize}

The protocol proceeds in four steps:
\begin{enumerate}
  \item Agent~A requests a trait evaluation of Agent~B via the runtime
        server.
  \item Agent~B polls the server for task requests, accepts the task, and
        receives a containerized executable, which, in our case, is the embedding
        model that produces diff vectors from mounted directories
        containing Agent~B's before and after skill files.
  \item Agent~B runs the executable locally on its own before/after skill
        files and submits the raw diff vectors to the runtime server.
  \item The runtime server runs a processor, which applies the trait
        vector to the diff vectors, computes the scalar trait score, and
        returns the result to Agent~A.
\end{enumerate}

This design has two key properties: agents are pure HTTP clients, so
neither exposes inbound endpoints, and execution is separated from
server-side scoring. The threat model below describes the resulting
protections.

\subsection{Trust and Continuity}
\label{sec:trust}

In our intended usage, Agent~A periodically requests trait evaluations
covering only the skill file updates since the previous assessment,
maintaining a running view of Agent~B's trait trajectory. Ensuring that
consecutive evaluations are anchored to the correct file versions can be
achieved through hash chaining: Agent~B includes a hash of each before
and after file, and Agent~A references the previous ``after'' hash in
its next request. A full specification is left to future work.

\paragraph{Threat model.}
The protocol is intended for cooperative deployments where Agent~A and
Agent~B have an existing relationship but should not share sensitive
files or scoring authority. A malicious Agent~A cannot directly inspect
Agent~B's files or choose arbitrary evaluation code: Agent~B receives a
digest-pinned evaluator from the trusted runtime server and submits only
diff vectors. A malicious Agent~B cannot simply report a favorable score,
since the server controls the trait vector and scoring processor. The
protocol does not address a known adversary that can substitute fake
skill trees or hide unassessed files; future hardening could add
repository-level commitments such as Merkle-tree commitments over the
assessed file set.

\subsection{Deployment}
\label{sec:deployment}

To validate the full system end-to-end, we deployed the executor and
processor containers described in \cref{sec:protocol} with a live runtime
server. The executor calls Qwen3-Embedding-8B on Google Vertex AI and
computes diff vectors from mounted skill file directories; the processor loads
the trained trait vector and computes scalar scores. Both containers are
stored in a registry and referenced by digest for reproducibility.

We deployed a Hermes~\cite{hermes2026} agent, a general-purpose agent
that we use as an engineering assistant, acting as both
Agent~A and Agent~B. A Hermes cron job checks for
skill files updated since the last successful evaluation. When updates
are found, the agent initiates the full protocol flow: it creates a task
request (as Agent~A), accepts it (as Agent~B), receives the executor
container image from the runtime server,
runs the container locally on its before/after skill directories, and
submits the resulting diff vectors. The runtime server then runs the
processor container and logs the trait diff scores along with run
metadata to a database. The agent (as Agent~A) polls until the result is
available.

We applied the data-seeking trait across all of the agent's
skills, which included both skills native to Hermes and skills we had
created for it. For newly created skills with no prior version, we
recommend manually reviewing the initial version to establish a baseline
trait value, since the embedding of an empty string may not occupy a
semantically meaningful region of the space; subsequent diffs then
proceed normally. In our deployment, the system flagged its largest
trait diff when substantial SSH and VM provisioning functionality was
added to a skill---an intuitive result that demonstrates practical
utility.

\section{Limitations and Future Work}
\label{sec:limitations}

\paragraph{Scope.}
We validate one trait, data-seeking, on 63 public ``before'' skills from
one repository. Extending to more traits, more diverse deployed-agent
skills, and correlations between traits is future work. Our core unit is
a single skill update; aggregating per-skill scores into agent-level risk
estimates is another practical next step, and we describe initial
heuristics in Appendix~\ref{sec:aggregation}.

\paragraph{Broader applicability.}
The approach rests on trust assumptions---the embedding model and the
runtime server are trusted intermediaries---which future integrity
measures such as HMAC verification of executor output and nonce binding
to prevent replay attacks could further strengthen.
As the ecosystem of interacting agents grows,
standardized mechanisms for agents to assess one another's properties
will be a prerequisite for scalable multi-agent trust.

\section*{Impact Statement}

This paper presents work toward enabling monitoring and evaluation of AI
agent behavior through their configuration files. By providing tools to
detect potentially dangerous trait changes such as increased
data-seeking behavior, this work aims to improve the safety and
trustworthiness of deployed agents. We note that the same methodology
could in principle be used by a malicious actor to craft edits that evade
detection; adversarial robustness of trait vectors is an important
direction for future work.

\bibliography{references}
\bibliographystyle{icml2026}

\appendix

\section{Aggregation Heuristics}
\label{sec:aggregation}

The core unit of our approach is a single update to a single skill;
however, one would like to roll up these updates into agent-level risk estimates at a point in
time. We use two aggregation heuristics.

First, for a single skill, we sum the trait diffs between consecutive
commits over time to obtain an absolute trait level at the current point
in time. To make this value actionable, we can monotonically map it to a
probability of the skill inducing unintended data-seeking behavior.
This mapping can be calibrated either through observed agent behavior or
top-down estimates.

Then, treating each skill independently, we view the agent-level risk as
the probability that \emph{some} skill causes a violation:
$R = 1 - \prod_i (1 - p_i)$, where $p_i$ is the per-skill risk
probability. As the number of skills grows, this product can drive $R$
toward~1 even when individual risks are low. To account for this, we
normalize by skill usage via a weighted geometric mean:
$R = 1 - \bigl(\prod_i (1-p_i)^{n_i}\bigr)^{1/N}$, where $n_i$ is the
invocation count for skill $i$ over a specified time period and $N = \sum_i n_i$, so that
infrequently used skills contribute proportionally less.

\end{document}